\documentclass{vgtc}                          % final (conference style)
\graphicspath{{figures/}{pictures/}{images/}{./}} % where to search for the images

\usepackage{times}                     % we use Times as the main font
\usepackage{tabu}                      % only used for the table example
\usepackage{booktabs}                  % only used for the table example
\usepackage{lipsum}                    % used to generate placeholder text
\usepackage{mwe}                       % used to generate placeholder figures

\usepackage{mathptmx}                  % use matching math font

\onlineid{0}

\vgtccategory{Research}

\vgtcinsertpkg

\title{\textsc{Diffusion Explorer}: Interactive Exploration of Diffusion Models}

\author{Alec Helbling\thanks{e-mail: alechelbling@gatech.edu}\\ %
        \scriptsize Georgia Tech %
\and Polo Chau\\ %
     \scriptsize Georgia Tech }

\teaser{
  \centering
  \includegraphics[width=0.98\linewidth]{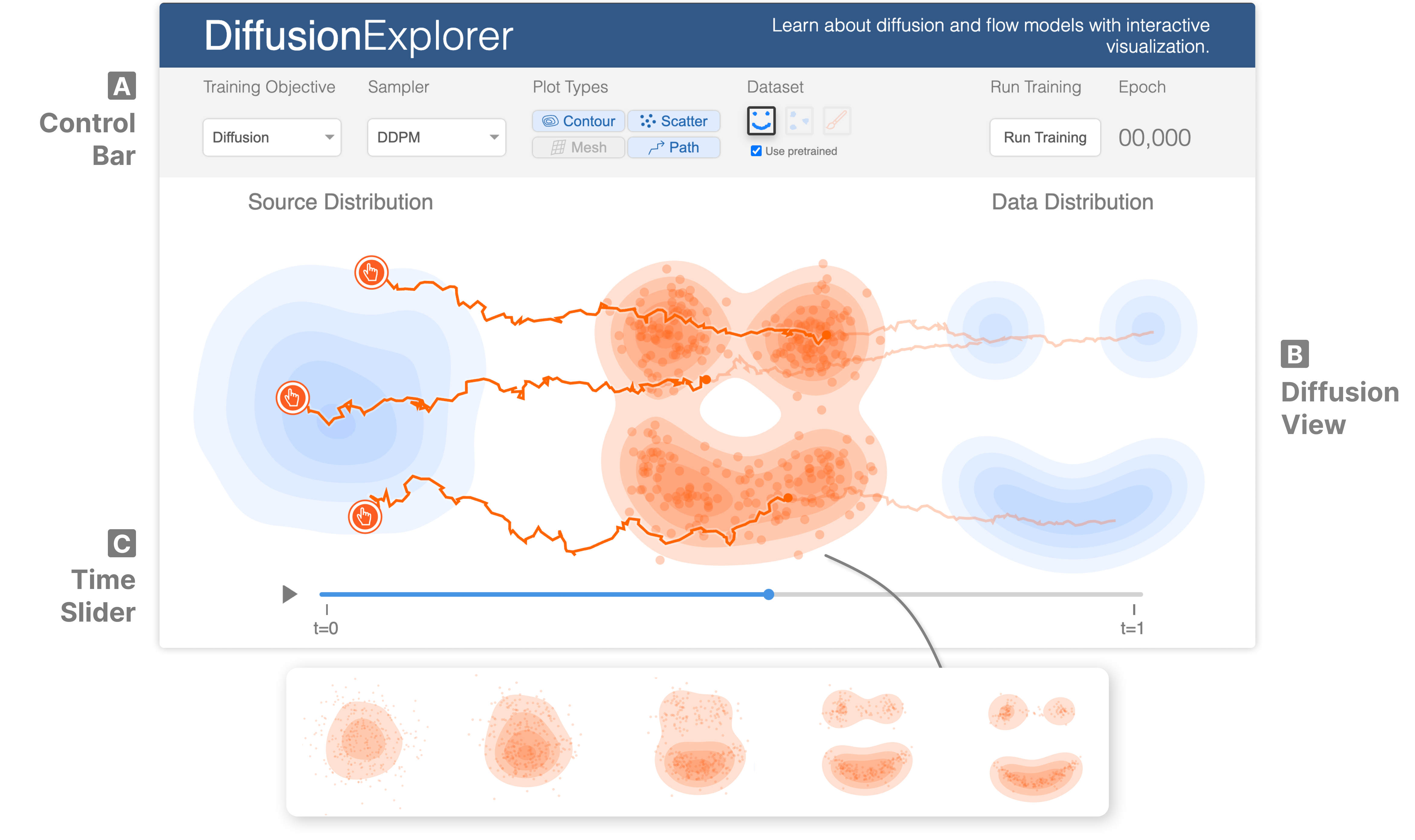}
  \vspace{-0.13in}
  \caption{With \textsc{DiffusionExplorer}, users can interactively train diffusion models and observe the temporal evolution of samples as they are transformed from a simple source distribution to a complex data distribution. \textbf{(A)} The control bar enables users to explore various model hyperparameters, plot types and datasets. \textbf{(B)} 
  The diffusion view of our interface helps users simultaneously learn about the dual nature of (1) the low-level stochasticity of individual samples by examining their trajectories; and (2) the smooth transformation of the high-level distribution through a contour plot visual summary. 
  \textbf{(C)} The time slider allows users to manually control an abstract notion of time, interpolating between the source and data distribution. Below the main interface we show the progression of samples starting out as a standard Gaussian distribution and eventually approximating a smiley face shape. }
  \label{fig:teaser}
}

\abstract{
    Diffusion models have been central to the development of recent image, video, and even text generation systems. They posses striking geometric properties that can be faithfully portrayed in low-dimensional settings. However, existing resources for explaining diffusion either require an advanced theoretical foundation or focus on their neural network architectures rather than their rich geometric properties. We introduce Diffusion Explorer, an interactive tool to explain the geometric properties of diffusion models. Users can train 2D diffusion models in the browser and observe the temporal dynamics of their sampling process. Diffusion Explorer leverages interactive animation, which has been shown to be a powerful tool for making engaging visualizations of dynamic systems, making it well suited to explaining diffusion models which represent stochastic processes that evolve over time. Diffusion Explorer is open source and a live demo is available at \url{alechelbling.com/Diffusion-Explorer}. 
} % end of abstract


\begin{document}


\firstsection{Introduction}

\maketitle

Diffusion models \cite{ddpm} 
have become one of the key model classes underpinning advancements in image, video, and text generation. 
A diffusion model learns to draw new samples of data by transforming random noise into arbitrarily complex data distributions like the natural images. 
However, despite their importance in the modern machine learning literature, there is a clear lack of accessible resources that effectively communicate the geometric properties of diffusion models to a novice audience. Existing resources either assume an advanced theoretical foundation \cite{nakkiran2024stepbystepdiffusionelementarytutorial} or focus on specific application settings like text-to-image generation and the nuances of their neural network architectures \cite{lee2024diffusionexplainervisualexplanation}. 

Animation has been shown to be a powerful technique for communicating the dynamic behavior of algorithms \cite{brownTechniquesAlgorithmAnimation1985}, effectively increasing learner engagement \cite{aminiHookedDataVideos2018, helbling2023manimmlcommunicatingmachinelearning}, and can be especially helpful for showing transitions between the states of a system \cite{heerAnimatedTransitionsStatistical2007}. Astonishingly, many of the rich geometric properties of diffusion models can be faithfully depicted in simple 2D settings; it is even possible to train diffusion models on 2D distributions using the exact same theoretical framework that scales to high dimensional modalities like images and video. 
\newpage
Filling in this gap, we contribute:
\begin{enumerate}
    \item \textbf{Diffusion Explorer, an interactive tool designed to explain the geometric properties of diffusion models to non-experts. } Diffusion Explorer leverages animation to depict the temporal evolution of data samples as they are transformed from simple source distributions to arbitrarily complex data distributions. Users can observe that path that individual samples take as well as see the high level transformation of the entire distribution of data. Users can also choose between different training objectives and sampling strategies, toggle different plot types, and even train their own diffusion models on hand drawn distributions in real time. 
    \item \textbf{A browser-based, open source implementation. } Diffusion Explorer runs completely on the front end, leveraging TensorFlow.js \cite{smilkov2019tensorflowjsmachinelearningweb} to draw samples from and train diffusion models in real time and D3.js \cite{2011-d3} to craft vector graphic animations. 
\end{enumerate}

\begin{figure}[t!]
    \centering
    \includegraphics[width=\linewidth]{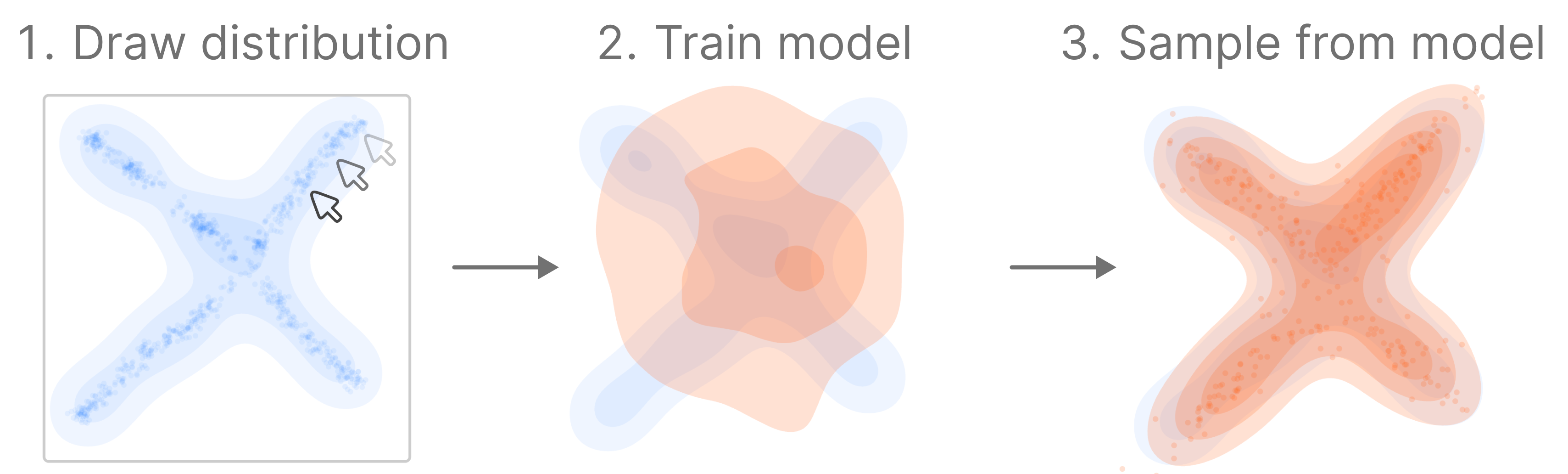}
    \caption{With \textsc{DiffusionExplorer}, users can (1) draw a custom distribution with their mouse, (2) train the model in browser, observing the evolution of the distribution as the model converges, and (3) sample from the trained model. }
    \label{fig:training}
\end{figure}

\section{System Design}

Diffusion Explorer has two key functionalities: (1) it shows how samples drawn from diffusion models evolve over time, and (2) allows users to draw custom data distributions and observe how the samples converge on the true data distribution. The interface of Diffusion Explorer has several key components. The control bar (A) allows users to select from different training objectives and sampling algorithms, comparing and contrasting their behavior. The diffusion view (B) is the central section of the interface, displaying the source, target, and current distributions. Finally, the time bar (C) allows users to manually control an abstract notion of time, where $t=0$ corresponds to the random noise distribution and $t=1$ corresponds to the data distribution. 

\subsection{Sampling from Pretrained Models } One of the key functionalities of Diffusion Explorer is to allow users to observe the transformation of samples from a source distribution to the data distribution. Samples are initially drawn from a Gaussian \textit{source distribution} and they are updated over time from $t=0$ to $t = 1$ until they are transformed to the \textit{data distribution}. It is worth noting that the convention in the diffusion literature is to use $t=1$ to correspond to the noise distribution and $t=0$ the data distribution, but we opted for the opposite convention for simplicity. 

Users can select locations in the source distribution and observe the trajectories of these samples over time as they are mapped from the source distribution to the data distribution (See Figure \ref{fig:teaser}). The stochastic nature of diffusion samples is evident from the abrupt movements of the samples. Diffusion Explorer also highlights the high level transformation of the entire distribution of samples over time using contour plots. This dichotomy between individual sample behavior and the distribution level properties of the model is a key property of diffusion models. 

This sampling process is done entirely in the browser user Tensorflow.js \cite{smilkov2019tensorflowjsmachinelearningweb}, and actually draws from real diffusion models trained on 2D distributions. Users can select various different training objectives, sampling algorithms, and plot types in the control bar. Diffusion Explorer supports two predefined datasets (a smiley face and three dots) that are accompanied by pre-trained models. The animations are primarily implemented in D3.js \cite{2011-d3}. 

\subsection{Training Models in Browser} 
In addition to allowing users to sample from pre-trained diffusion models, they can also train models on their own hand-drawn datasets. After selecting the paint-brush icon in the dataset menu of the control bar users are prompted to draw a custom distribution shape and a diffusion model is trained to draw samples from this distribution (see Figure \ref{fig:training}). Training is done completely in the browser using Tensorflow.js and it takes on the order of 10 second for the model to converge to a reasonable approximation of the data distribution. Diffusion Explorer shows samples drawn from the trained models after each epoch, which gives the user insight into how the model evolves during training. Users can then draw samples from their custom trained model and see how the samples evolve over time. 

\section{Initial Feedback and Ongoing Work}

We are excited by the positive feedback we have gotten from the research community. At the time of writing, our GitHub repository has received over 750 stars, thousands of page views, and significant engagement from the community on social media. Some people have already reached out and asked if they can incorporate our tool into their machine learning undergraduate courses. 

We plan to make some key extensions of our work like adding additional training objectives, sampling algorithms, and plot types. Additionally, we plan to make a complementary narrative guided interactive article that explains some of the theoretical concepts underpinning diffusion models. We aim to leverage a similar visualization style and perform an extensive user study comparing the efficacy of our tool and our planned interactive article as a learning tool. Our hypothesize is that our open ended tool will become much more useful as an educational tool when combined with contextually relevant details about how these models work. 

\bibliographystyle{abbrv-doi}

\bibliography{template}
\end{document}